\documentclass{article}

\usepackage{arxiv}

\usepackage{amsmath,amssymb,amsfonts}
\usepackage{algorithmic}
\usepackage{graphicx}
\usepackage{bbm}
\usepackage{textcomp}
\usepackage{algorithmic}
\usepackage{algorithm}
\usepackage{booktabs}
\usepackage{multirow}
\usepackage{xspace}
\usepackage{url}
\usepackage{subcaption}
\usepackage[justification=centering]{caption}
\usepackage[symbol]{footmisc}

\usepackage{scrextend}

\usepackage[utf8]{inputenc}  
\usepackage[T1]{fontenc}    
\usepackage{hyperref}       
\usepackage{url}            
\usepackage{booktabs}       
\usepackage{amsfonts}       
\usepackage{nicefrac}       
\usepackage{microtype}      
\usepackage{lipsum}		
\usepackage{graphicx}
\usepackage{natbib}
\usepackage{doi}
\usepackage{authblk}

\usepackage{hyperref}

\def\BibTeX{{\rm B\kern-.05em{\sc i\kern-.025em b}\kern-.08em T\kern-.1667em\lower.7ex\hbox{E}\kern-.125emX}}

\title{A simple self-supervised ECG representation learning method via manipulated temporal--spatial reverse detection}

\author[ ]{Wenrui Zhang\textsuperscript{1} , Shijia Geng\textsuperscript{2} , Shenda Hong\textsuperscript{3,4}\footnote[1*]{} }

\affil[1]{Department of Mathematics, National University of Singapore, Singapore, 119077, Singapore}
\affil[2]{HeartVoice Medical Technology, Hefei, 230027, China}
\affil[3]{National Institute of Health Data Science, Peking University, Beijing, 100191, China}
\affil[4]{Institute of Medical Technology, Health Science Center of Peking University, Beijing, 100191, China}

\renewcommand{\shorttitle}{\textit{arXiv} Template}

\begin{document}
\maketitle
\renewcommand{\thefootnote}{\fnsymbol{footnote}}
\footnotetext[1]{Corresponding authors: Shenda Hong, Email: hongshenda@pku.edu.cn}

\begin{abstract}
Learning representations from electrocardiogram (ECG) signals can serve as a fundamental step for different machine learning-based ECG tasks. In order to extract general ECG representations that can be adapted to various downstream tasks, the learning process needs to be based on a general ECG-related task which can be achieved through self-supervised learning (SSL). However, existing SSL approaches either fail to provide satisfactory ECG representations or require too much effort to construct the learning data. In this paper, we propose the \textbf{T-S reverse} detection, a simple yet effective self-supervised approach to learn ECG representations. Inspired by the temporal and spatial characteristics of ECG signals, we flip the original signals horizontally (temporal reverse), vertically (spatial reverse), and both horizontally and vertically (temporal--spatial reverse). Learning is then done by classifying four types of signals including the original one. 
To verify the effectiveness of the proposed method, we perform a downstream task to detect atrial fibrillation (AF) which is one of the most common ECG tasks. The results show that the ECG representations learned with our method achieve remarkable performance.
Furthermore, after exploring the representation feature space and investigating salient ECG locations, we conclude that the temporal reverse is more effective for learning ECG representations than the spatial reverse.

\end{abstract}

\keywords{
ECG \and Self-Supervised Learning \and Representation Learning \and Deep Learning}

\section{Introduction}
Machine learning, especially deep learning has been widely used for different ECG tasks \cite{hong2020opportunities}, such as disease detection \cite{rahul2022artificial, han2022detecting, ertuugrul2021automatic,tong2021locating}, sleep staging \cite{zhao2021dual, werth2020deep}, biometric human identification \cite{hong2020cardioid,zhang2021human}, and denoising \cite{rasti2022deep}. Good general ECG representations can provide a common intermediate 
abstraction layer for different ECG tasks and speed up the training processes. Many studies have begun to explore ECG representation learning \cite{liu2021self,sarkar2020self,kiyasseh2021clocs,lan2021intra}. In order to learn general ECG representations, the task used for learning needs to be a general ECG-related task, which can be achieved by self-supervised learning (SSL). SSL uses constructed inherent supervisory information to learn the underlying representations, which avoids the cost of manual data annotation and provides a flexible training paradigm \cite{liu2021self}. For example, in order to learn general language representations, the BERT model performs Cloze tests based on randomly masked unlabeled text \cite{devlin2018bert}. 
The following two types of methods are often used in SSL.
\begin{itemize}
    \item \textbf{Reconstruction}: reconstruction-based approaches treat representation learning as a ``deconstruction" process, and aim to rebuild the original data from the representations. The assumption is that good representations should reserve enough information that can reconstruct the original data. Such approaches do not require any labelling or construction of learning data, and obtain representations through dimension reduction methods such as principal component analysis \cite{wold1987principal} (PCA) and autoencoder \cite{bourlard1988auto} (AE) which are widely used for ECG data \cite{dasan2021novel, porumb2020nocturnal,kuznetsov2021interpretable}. Although reconstruction-based approaches are easy to implement, the performances of downstream tasks based on representations learned from these methods are often unsatisfactory. 
    
    \item \textbf{Contrastive learning (CL)}: CL-based methods have drawn much attention in recent years, and have been applied in learning ECG representations \cite{sarkar2020self,kiyasseh2021clocs,lan2021intra}. The basic idea is to create a model for judging similarities and differences, assuming that good representations can facilitate the judgement. The ECG representations learned from CL show great performances on downstream tasks \cite{liu2021self}. However, the framework of CL is complex, and it is difficult and sometimes impossible to construct similar and dissimilar pairs for training.
\end{itemize}

To address the problems of the above methods, we design an SSL task that not only generates effective representations for downstream ECG tasks, but also saves time and effort in constructing learning data. Inspired by the ECG temporal and spatial characteristics, we manipulate ECG segments with three simple transformations: flipping horizontally (corresponding to temporal reverse), vertically (corresponding to spatial reverse), and both horizontally and vertically (corresponding to temporal-spatial reverse), and then train the representation learning model to classify these four types of segments. 

We conduct atrial fibrillation (AF) detection as a downstream task to evaluate our learned representations. Experimental results show that ECG representation learning via T-S reverse detection outperforms other baseline methods by a large margin. We also find that the temporal reverse helps learn better representations than the spatial reverse. 

\section{Materials and methods}

\subsection{Dataset}
We utilize the publicly available and widely used dataset from PhysioNet/Computing in Cardiology Challenge 2017 (CinC 2017) \cite{clifford2017af}. The data is collected and band-pass filtered by the AliveCor device. The sampling rate of all ECG signals is 300 Hz, and the time length ranges from 9 s to over 60 s. The whole training set contains 8528 single-lead ECG recordings with four classes. We select ECG signals with ``Normal" and ``AF" labels, and report the statistics in Table \ref{tb:db}. 

\begin{table}
    \centering
    \caption{Statistics of our data. SD means standard deviation.}
    \label{tb:db}
    \begin{tabular}{c|c|c|c|c|c|c}
    \toprule
        \multirow{2}{*}{Type} & \multirow{2}{*}{\# of recordings} & \multicolumn{5}{c}{Time length (s)} \\\cline{3-7}
         & & Mean & SD & Max &  Median & Min \\\midrule
         Normal&5075&31.9&10.0&61.0&30&9.0 \\\cline{1-7}
         AF&758&31.6&12.5&60&30&10.0\\\bottomrule
    \end{tabular}
    
\end{table}

We perform data preprocessing following two steps: (1) we split all recordings into 10-second segments $x \in \mathbb{R}^{L}, L=3000$ using a sliding window with a 5-second stride. The segments that are shorter than 3000 data points (10 s $\times$ 300 Hz) are discarded. (2) we normalize each signal to [0, 1] based on the equation: $x_{normalized} = \frac{x - \min(x)}{\max(x) - \min(x)}$. 

\subsection{Overall framework}
\begin{figure}
    \centering
    \includegraphics[width=\linewidth]{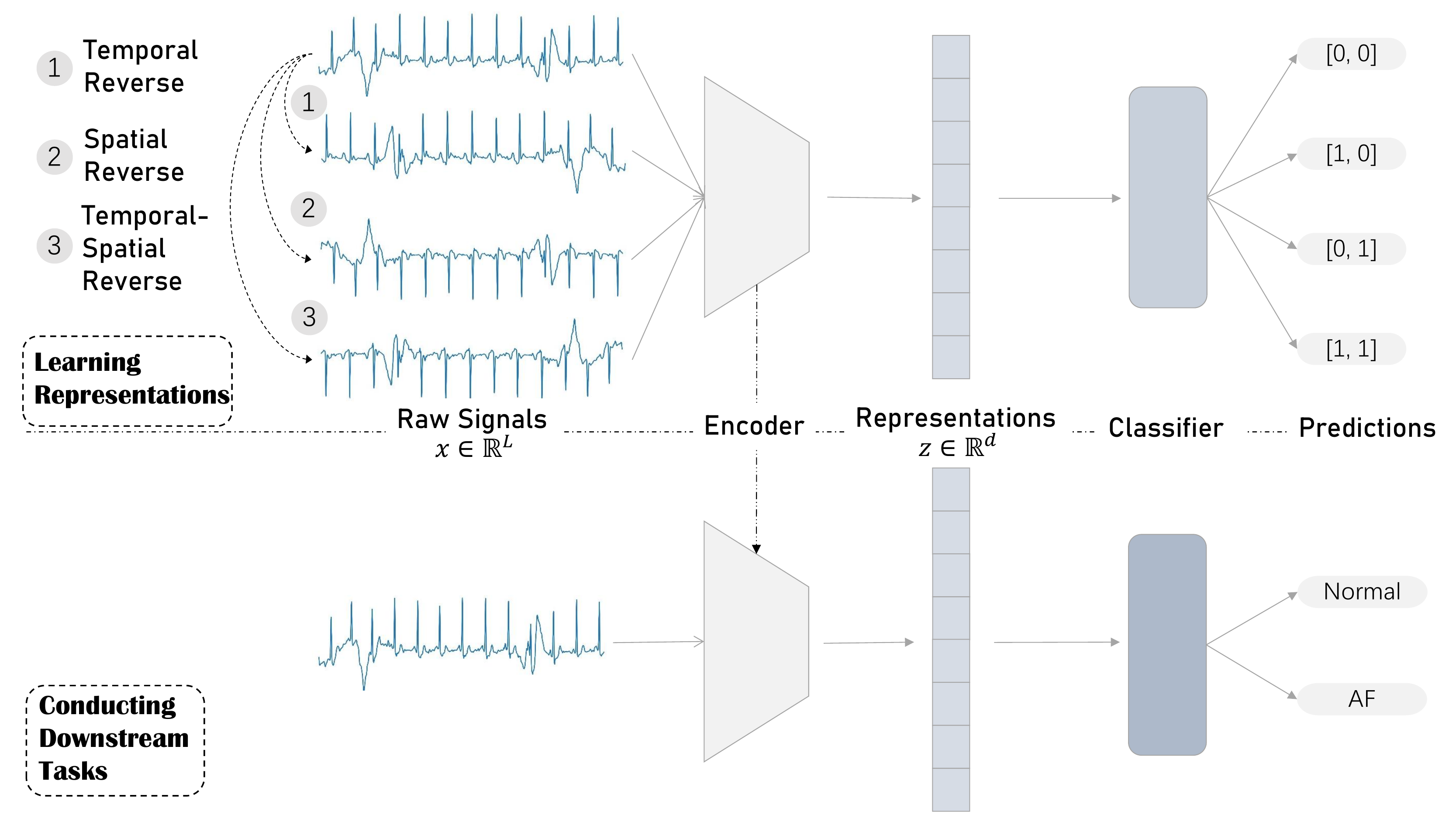}
    \caption{Overview of the two-stage framework with our method. In the first stage (learning representations), the encoder is trained by classifying different reversed and original ECG segments. In the second stage (conducting downstream tasks), the trained encoder is transferred to extract ECG representations, and a new classifier is fine-tuned to perform the specific downstream task, which is distinguishing between  Normal and AF ECG segments in this experiment.}
    \label{fig:overview}
\end{figure}

The overall two-stage framework is shown in Figure \ref{fig:overview}. The first stage is to learn representations based on our T-S reverse detection method, which is also called the pre-trainng stage. The learning model consists of two parts: encoder and classifier. The encoder produces the representations, and the classifier is for conducting the T-S reverse detection task during the learning process. After the learning is finished, the trained encoder is transferred to the second stage for various downstream tasks. We only show our experimental AF detection downstream task in Figure \ref{fig:overview}.

\subsection{Pre-training based on manipulated T-S reverse detection}
\begin{figure}
    \centering
    \includegraphics[width=\textwidth]{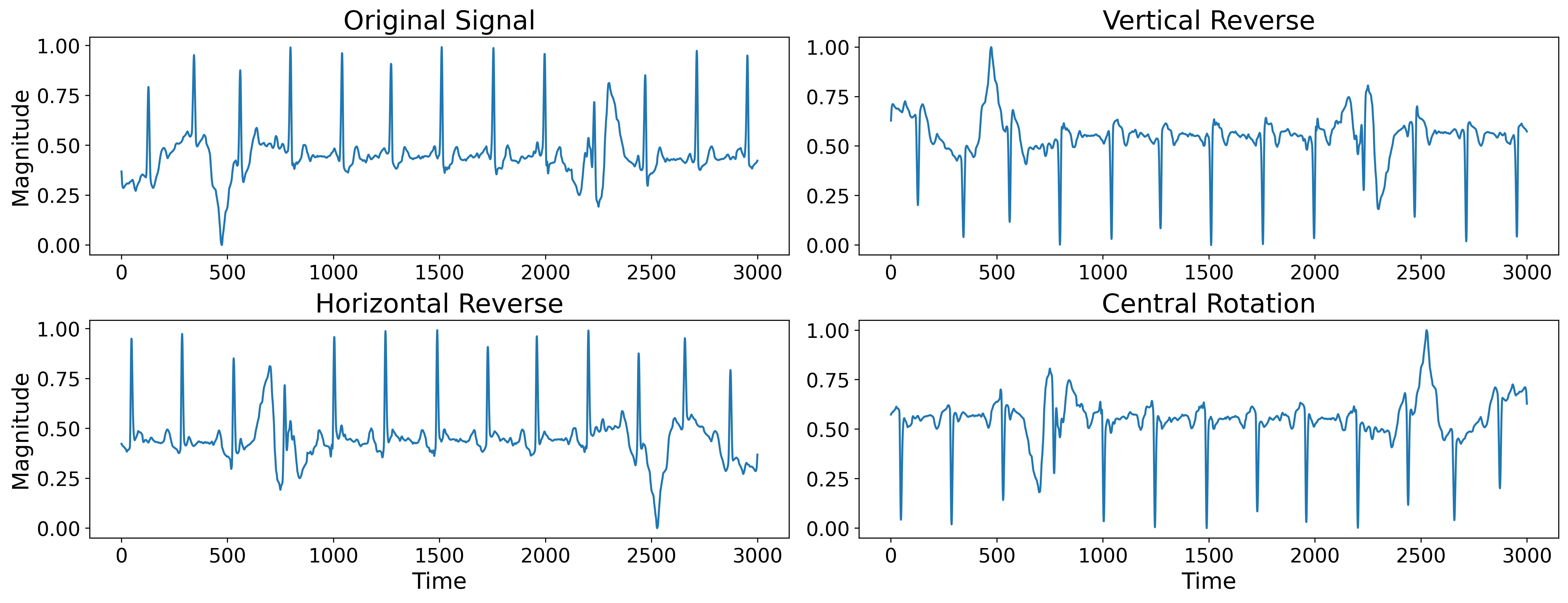}
    \caption{Original ECG segment and three types of reversed segments.}
    \label{fig:transform}
\end{figure}

To provide supervisory information to unlabelled data, we manipulate the original ECG signals with three types of reverse methods (T-S reverse manipulations) which include temporal reverse (flipping horizontally), spatial reverse (flipping vertically), and temporal-spatial reverse (flipping both horizontally and vertically). The examples are illustrated in Figure \ref{fig:transform}, and the details of the reverse manipulations are listed as follows.
\begin{itemize}
    \item \textbf{Temporal reverse}: we regard flipping horizontally as temporal reverse, because it changes the temporal information of original signals. We hypothesize that by discriminating horizontally reversed signals from original signals, the model can learn sequential information of ECG signals. 
    We let $x'\text{[i]} = x[3000-i+1], \text{i}=1,2,\dots,3000 $
    where $x$ is a 10-second segment and 3000 is the segment length.
    
    \item \textbf{Spatial reverse}: we see flipping vertically as a method changing the spatial information of original signals, so we call it spatial reverse. We aim to enable the model to ``observe" the spatial difference, so that the model can capture the information on vertical direction of ECG signals. The implementation can be divided as 2 steps. First, we multiply each value of signals by -1, so all signals are vertically reversed and in a range of [-1, 0]. Second, we re-normalize the reversed signals, and all signals are reversed vertically and moved to [0, 1]. 

    \item \textbf{Temporal-Spatial reverse}: temporal-spatial reverse is a combination of temporal and spatial reverse. By performing temporal-spatial reverse, we can enlarge the amount of training data. When performing temporal-spatial reverse, we only need to make a horizontal reverse on the vertically reversed segments. 
    
\end{itemize}

After the reverse manipulations, we mark the segments using one-hot encoding. The original segments are marked as [0, 0], the temporal reverse segments are marked as [0, 1], the spatial reverse segments are marked as [1, 0], and the temporal-spatially reverse segments are marked as [1, 1]. we train the representation learning model to classify the different reversed and original segments. We hypothesize that by detecting the four classes of segments, the model can learn both temporal and spatial information of ECG signals, and extract features (such as QRS Complex) that are essential for downstream ECG tasks. 

The learning model is a 32-layer deep neural network modified from the ResNeXt network architecture \cite{ResNext}, which is designed following a practical network design paradigm \cite{radosavovic2020designing} and achieves state-of-the-art performances on ECG-related tasks \cite{Hong_2019,hong2020holmes}. In our experiment, the first 31 layers serve as an encoder that maps segments into representations, and the last layer is a fully-connected neural network layer that serves as a classifier for the pre-training task.

\subsection{Fine-tuning for the downstream task}
After T-S reverse detection, we obtain a model that acquires the ability to identify the original ECG segment from the reverse replica. The model has ``understood" what a real ECG segment is, and is able to abstract the ECG representations. To transfer the knowledge and apply it to other tasks, we take the encoder part that maps an ECG segment $x$ (without reverse manipulations) to its representation $z \in \mathbb{R} ^ d$ (where $d$ is the dimension of the learned representation). We apply a new classifier (a fully-connected neural network layer) on $z$, and train the classifier to conduct the AF detection downstream task based on the data labels (Normal and AF). The input of the classifier is $z$ and the output is 0 or 1 which represents Normal or AF. The training process at this stage is known as fine-tuning, as it usually only requires small adjustments to the pre-trained model.

\subsection{Model interpretation}
To further understand how the knowledge learned from the T-S reverse detection works, we employ layer-wise relevance propagation (LRP) \cite{bach2015pixel} to interpret the overall downstream model (including the transferred encoder and the AF classifier). LRP has been applied to many tasks related to physiological signals, such as electroencephalography (EEG) classification \cite{sturm2016interpretable} and ECG classification \cite{banluesombatkul2020metasleeplearner}. 
LRP can produce a heatmap that indicates how the input data points correlate with a neural network's final decision. The output value is backprojected layer to layer onto the inputs, and provides one relevance score ($R$ score) to each data point of the inputs. The higher the $R$ score is, the larger contribution the data point provides to the final decision of the model.

\section{Experimental Setup}

\subsection{Baseline methods}
For comparison with our T-S reverse detection method we conduct four representation learning baseline methods. Details of each baseline method are as follows.
\begin{itemize}
    \item \textbf{Random Projection (RP)}: RP maps the original input space to the target space through a randomly generated matrix. The elements of the random matrix are drawn from a normal distribution $\mathcal{N}(0, \frac{1}{dim})$, where $dim$ is the target dimension.

    \item \textbf{PCA}: PCA can be used to decompose a high-dimensional dataset. High-dimensional data is projected to a set of successive orthogonal components that contribute most to the variance.

    \item \textbf{AE}: AE is composed of two parts: the encoder and the decoder. The encoder is to map the original input $x$ in data space to $z$ in feature space, and the decoder is to reproduce $x'$ (a reconstruction of $x$) by mapping $z$ back to data space. AE is trained by minimizing the reconstruction loss $\mathcal{L} = \sum_{i=1}^N(x_i-x'_i)^2$, where $N$ is the dimension of original data. In our experiment, we conduct a simple implementation of AE with a ResNet-18 network \cite{he2016deep} as the encoder, and the decoder has the symmetric structure. 
    
    \item \textbf{SimCLR \cite{chen2020simple}}: SimCLR is a contrastive learning method. We conduct two types of data augmentation methods on raw signals, which are permutation and adding noise. The same segments and their replica with the two augmentation methods are seen as positive pairs, and any two different segments (with either same or different augmentation methods) are regarded as negative pairs. The loss function of each positive pair (i,j) is $\mathcal{L}=-\mathbb{E}[\frac{e^{sim(x_i, x_j)/\tau}}{\sum_{k\neq i}e^{sim(x_i, x_k)/\tau}}]$, where sim(x, y) represents cosine similarity $\frac{x^T \cdot y}{||x||\cdot||y||}$. We construct $N$ positive pairs from a batch of $N$ segments, and calculate the average of $N$ losses and minimize it. The model trained with SimCLR has the same architecture as the model trained to detect T-S reverse.
    
\end{itemize}

\subsection{Ablation study}
To evaluate the performance of temporal reverse detection and spatial reverse detection respectively, we conduct the ablation study as follows:
\begin{itemize}
    \item \textbf{Spatial reverse only detection}: the spatial reverse only detection task is a binary classification task to distinguish spatially (vertically) reversed segments from original segments, without regarding temporal reverse. The original signals are labeled as 0, and spatially reversed signals are labeled as 1.
    \item \textbf{Temporal reverse only detection}: the temporal reverse only detection task is a binary classification task to detect the temporally (horizontally) reversed segments. The original signals are labeled as 0, and temporally reversed signals are labeled as 1. 

\end{itemize}

\subsection{Different experimental settings}
To investigate the effect of ECG representation dimensions, we fix the number of fine-tuning segments as 2000, and test different dimensions set to 64, 128 and 256, respectively. 
In addition, using different amounts of training data (50, 100, 200, 500, 1000, 2000, and 20,000 ECG segments), we compare the performances of fine-tuning (based on ECG representations from T-S reverse detection) with the performances of the downstream model trained from scratch. Although data for fine-tuning is usually much less than data for pre-training, we evaluate our method when there is sufficient fine-tuning data, and conduct one of the tests with 20,000 ECG segments which is almost the entire training set.

We also verify the effect of different amounts of training data on ablation experiments.

\subsection{Evaluation metrics}
The performance metrics we select to evaluate the performance of different combinations are the receiver operating characteristics area under the curve (AUC), accuracy, sensitivity and specificity.
The calculation of the metrics are based on the following four terms.
\begin{itemize}
    \item True positive (TP): number of samples labeled positive and predicted positive.
    \item True negative (TN): number of samples labeled negative and predicted negative.
    \item False positive (FP): number of samples labeled negative but predicted positive.
    \item False negative (FN): number of samples labeled positive but predicted negative.
\end{itemize}

The x-axis of the ROC curve is the false positive rate ($\frac{FP}{FP+TN}$), and the y-axis is the true positive rate ($\frac{TP}{TP+FN}$). The AUC can measure the performance of a binary classifier with various discrimination thresholds. It can be regarded as the probability that a positive example ranks higher than a negative example \cite{AUC}. Accuracy is the ratio of the number of truly classified samples to the number of samples: $\text{Accuracy} = \frac{\text{TP+TN}}{\text{TP+TN+FP+FN}}$. Sensitivity (also called recall) is the correct classification rate only regarding positive samples: $\text{Sensitivity} = \frac{\text{TP}}{\text{TP+FN}}$. Specificity is the correct classification rate only regarding negative samples: $\text{Specificity} = \frac{\text{TN}}{\text{TN+FP}}$. For accuracy, specificity and sensitivity, we obtain the threshold for classifying labels by maximizing the geometric mean value \cite{gmean} G-mean: $\text{G-mean} = \sqrt{\text{Sensitivity} \times \text{Specificity}}$. 

\section{Results}

\subsection{Results of baselines and ablation study}

\begin{figure}
    \centering
    \includegraphics[width=\linewidth]{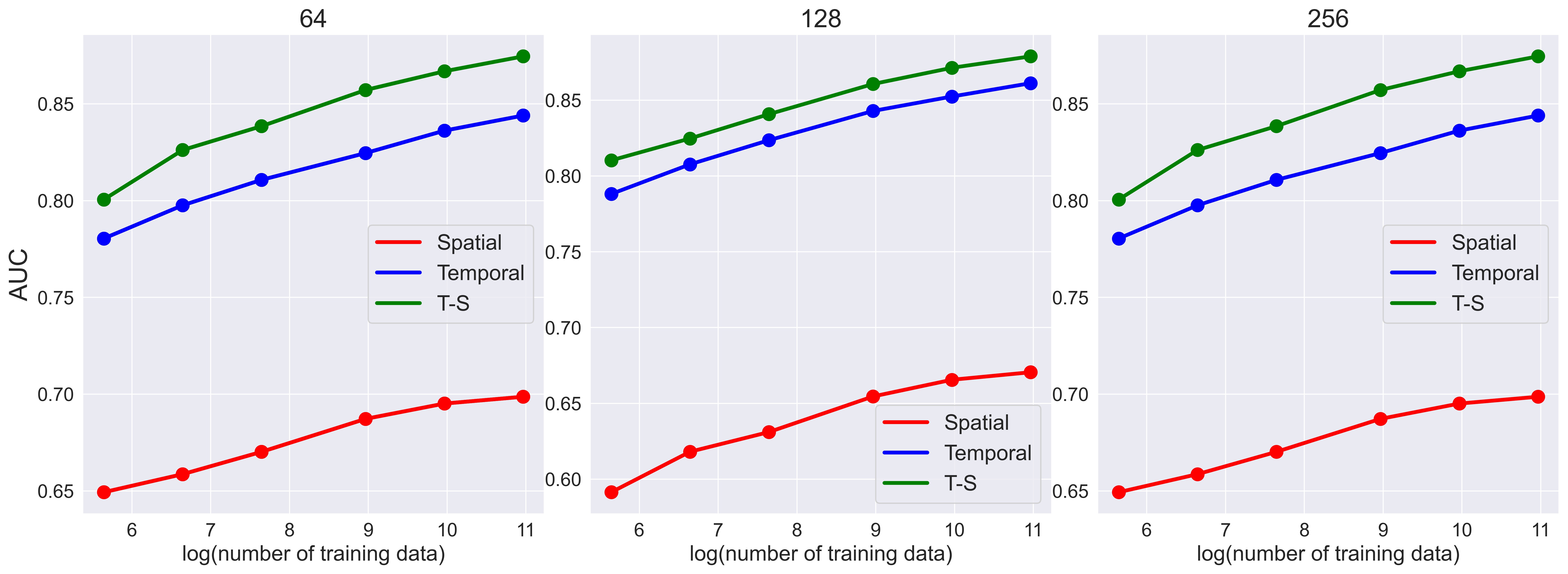}
    \caption{AUCs for 64, 128 and 256 representation dimensions with different numbers of training data.}
    \label{fig:auc}
\end{figure}

Table \ref{tab:results} shows the four adopted metrics of baseline methods (the first four rows of each section), ablation experiments (the next two rows), and our T-S reverse detection method (the last row). The results are obtained by AF detection using 2000 ECG segments (1000 segments per class). 
We can see that our T-S reverse detection method outperforms baseline methods on all metrics. For example, in terms of AUC, T-S achieves 35.8\%, 30.5\% and 25.6\% higher than the highest baseline when the dimensions of representations are 64, 128 and 256 respectively. For other metrics, the scores of baselines are below 0.7 most of the time, but the scores of T-S reverse are always above 0.8. As for the ablation study (see both Table \ref{tab:results} and Figure \ref{fig:auc}), we find that the spatial reverse detection is a relatively poor task for learning ECG representations, with significantly lower metrics than the temporal reverse detection. The performance of the temporal reverse method is similar to and sometimes even exceeds that of our T-S reverse method. It indicates that the temporal reverse contributes more to the performance of the T-S reverse detection method. 

Across different representation dimensions, we find that dimensions 128 and 256 outperform dimension 64, suggesting that dimension 128 may be a balance between performance and computational efficiency. Figure \ref{fig:tsne} supports this inference and shows that 128-dimension provides good performances even with a small amount of training data, while temporal reverse detection performs better when the representation dimension is 128.

We also compare the performances of our T-S reverse detection method and the model trained from scratch based on different amount of training data (50, 100, 200, 500, 1000, 2000, and 20,000 ECG segments). The results are shown in Table \ref{tab:result_ft}, and we can see that our method outperforms the model trained from scratch for all the different amounts of training data. It indicates that pre-training with T-S reverse detection helps improve the performance not only when the data is limited but also when data is sufficient.

\begin{table}
    \centering
    \caption{The results of baselines and our methods (including ablation experiments) trained with 2000 segments with different representation dimensions.}
    \label{tab:results}
    \begin{tabular}{c|c|c|c|c|c}
    \toprule
        Dimension & Method & AUC &  Sensitivity & Specificity & Accuracy  \\ \midrule
        \multirow{7}{*}{64}& RP & 0.536 $\pm$ 0.012 & 0.516 $\pm$ 0.035 & 0.555 $\pm$ 0.042 & 0.551 $\pm$ 0.032 \\
        &PCA & 0.597 $\pm$ 0.009 & 0.567 $\pm$ 0.032 & 0.584 $\pm$ 0.030 & 0.582 $\pm$ 0.023 \\
        &AE & 0.658 $\pm$ 0.015 & 0.594 $\pm$ 0.032 & 0.652 $\pm$ 0.028 & 0.645 $\pm$ 0.022\\
        &SimCLR & 0.662 $\pm$ 0.009 & 0.576 $\pm$ 0.038 & 0.664 $\pm$ 0.045 & 0.652 $\pm$ 0.036 \\
        &Spatial & 0.686 $\pm$ 0.009 & 0.577 $\pm$ 0.041 & 0.705 $\pm$ 0.042 & 0.689 $\pm$ 0.032\\
        &Temporal & 0.897 $\pm$ 0.006 & \textbf{0.831 $\pm$ 0.013} & 0.811 $\pm$ 0.014 & 0.814 $\pm$ 0.011\\
        &T-S & \textbf{0.899 $\pm$ 0.006} & 0.804 $\pm$ 0.013 & \textbf{0.832 $\pm$ 0.013} & \textbf{0.829 $\pm$ 0.010}\\
        
        \midrule
        
        \multirow{7}{*}{128}& RP & 0.551 $\pm$ 0.015 & 0.528 $\pm$ 0.029 & 0.556 $\pm$ 0.030 & 0.552 $\pm$ 0.024 \\
        &PCA & 0.586 $\pm$ 0.009 & 0.556 $\pm$ 0.021 & 0.581 $\pm$ 0.023 & 0.577 $\pm$ 0.018\\
        &AE & 0.662 $\pm$ 0.017 & 0.594 $\pm$ 0.037 & 0.660 $\pm$ 0.040 & 0.651 $\pm$ 0.032\\
        &SimCLR & 0.696 $\pm$ 0.008 & 0.597 $\pm$ 0.039 & 0.700 $\pm$ 0.044 & 0.687 $\pm$ 0.034\\

        &Spatial & 0.697 $\pm$ 0.010 & 0.601 $\pm$ 0.023 & 0.705 $\pm$ 0.025 & 0.693 $\pm$ 0.019\\
        &Temporal & 0.905 $\pm$ 0.005 & 0.821 $\pm$ 0.020 & 0.829 $\pm$ 0.023 & 0.828 $\pm$ 0.018\\
        &T-S & \textbf{0.908 $\pm$ 0.005} & \textbf{0.822 $\pm$ 0.018} & \textbf{0.836 $\pm$ 0.015} & \textbf{0.834 $\pm$ 0.011}\\
        
        \midrule
        
        \multirow{7}{*}{256}& RP & 0.553 $\pm$ 0.015 & 0.535 $\pm$ 0.039 & 0.552 $\pm$ 0.034 & 0.550 $\pm$ 0.026\\
        &PCA & 0.554 $\pm$ 0.012 & 0.521 $ \pm$ 0.027 & 0.566 $\pm$ 0.025 & 0.560 $\pm$ 0.019\\
        &AE & 0.655 $\pm$ 0.013 & 0.591 $\pm$ 0.034 & 0.647 $\pm$ 0.029 & 0.651 $\pm$ 0.022\\
        &SimCLR & 0.723 $\pm$ 0.007 & 0.628 $\pm$ 0.039 & 0.708 $\pm$ 0.046 & 0.696 $\pm$ 0.037\\

        &Spatial & 0.721 $\pm$ 0.010 & 0.630 $\pm$ 0.038 & 0.709 $\pm$ 0.038 & 0.700 $\pm$ 0.030\\
        &Temporal & 0.904 $\pm$ 0.009 & 0.810 $\pm$ 0.020 & \textbf{0.838 $\pm$ 0.019} & \textbf{0.834 $\pm$ 0.015}\\
        &T-S & \textbf{0.908 $\pm$ 0.008} & \textbf{0.830 $\pm$ 0.025} & 0.831 $\pm$ 0.023 & 0.831 $\pm$ 0.017\\
        \bottomrule
    \end{tabular}
    
\end{table}
\subsection{Data distributions in feature spaces}

\begin{table}
    \centering
    \caption{AUCs for the model trained from scratch and for the model pre-trained using T-S reverse detection with different numbers of training data.}
    \label{tab:result_ft}
    \begin{tabular}{c|c|c|c|c|c|c|c}
        \toprule
         \# of training segments & 50 & 100 & 200 & 500 & 1000 & 2000 & 20,000\\ \midrule
        From scratch (AUC) & 0.662 & 0.735 & 0.836 & 0.917 & 0.932 & 0.933 & 0.963\\ 
        T-S (AUC) & 0.873 & 0.906 & 0.936 & 0.956 & 0.966 & 0.973 & 0.981\\
        \bottomrule
    \end{tabular}
    
\end{table}

\begin{figure}
    \centering
    \includegraphics[width=\linewidth]{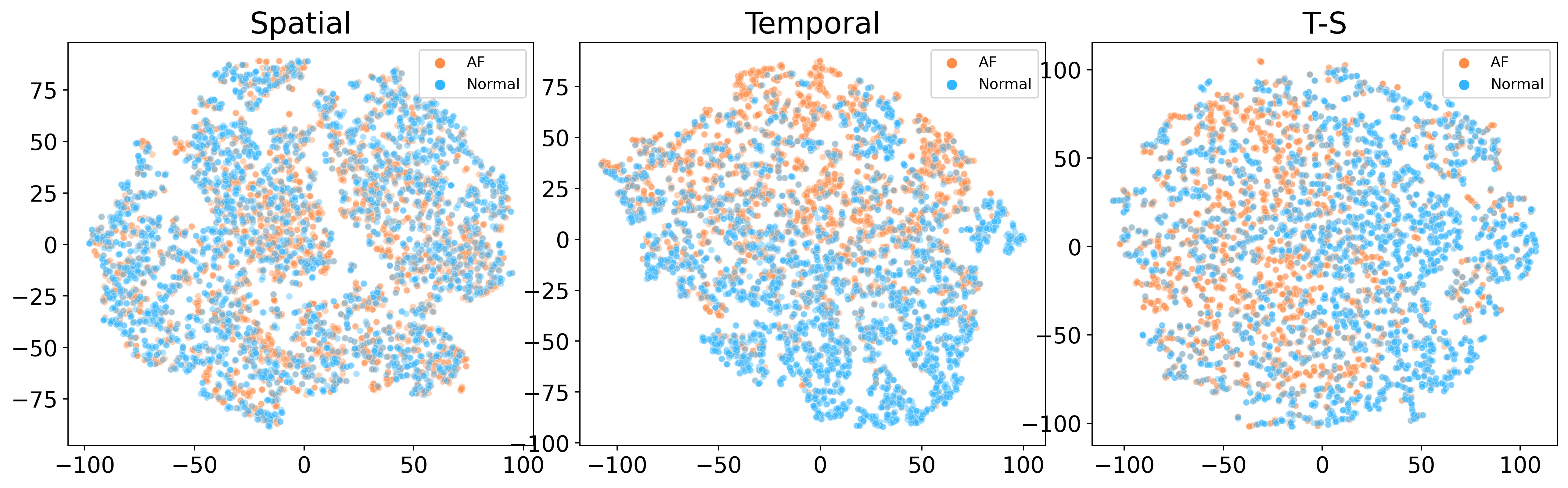}
    \caption{Data distributions in three feature spaces projected on 2-D planes. Yellow points represent label AF and blue points represent label Normal. We can see that in temporal and T-S feature spaces, features with the same label are more likely to be closer.}
    \label{fig:tsne}
\end{figure}
\begin{figure}
    \centering
    \includegraphics[width=\linewidth]{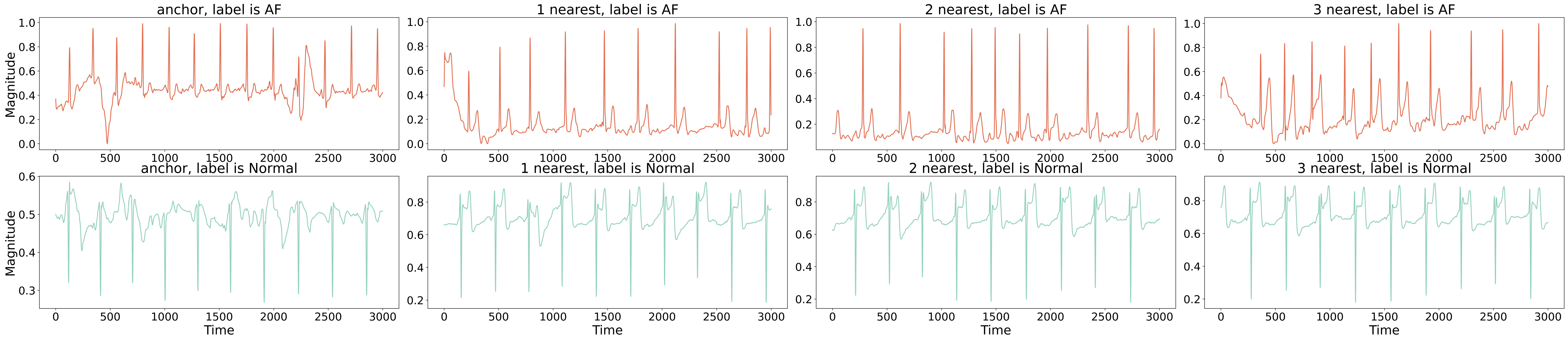}
    \caption{The segment examples with their top 3-nearest segments in the T-S feature space.}
    \label{fig:nearest}
\end{figure}

After pre-trained with one type of reverse detection, the model is able to project raw ECG segments into the corresponding latent representation space which is also known as the feature space. In Figure \ref{fig:tsne} (only dimension 64 is demonstrated), the first plot represents the feature space for the spatial reverse (spatial feature space), and we can see that yellow points corresponding to AF ECG segment and blue points corresponding to Normal ECG segments distribute evenly. It indicates that the model trained with spatial reverse cannot distinguish between AF and normal ECG segments. By contrast, the yellow AF points and the blue Normal points in the second and third plots are more discernible. From the second figure which represents the feature space for the temporal reverse (temporal feature space), we see more yellow points in the upper region and more blue points in the lower region. In the third figure that represents the feature space for the T-S reverse (T-S feature space), yellow points tend to be on the left and blue points tend to be on the right. The plots illustrate that models trained to classify temporal reverse and classify T-S reverse have better ``understandings" about ECG segments, and can provide more informative representations to distinguish AF from Normal.

We also conduct two case studies to investigate ECG segments with close distances in the T-S feature space. We randomly select 500 ECG segments with label 0 (Normal) and 500 ECG segments with label 1 (AF), and obtain the corresponding representations with the T-S reverse detection method. Then, we find the top-3 nearest representations for each selected segment, and calculate the mean of the labels. For example, if the labels corresponding to the top-3 nearest representations for one segment are 1, 1 and 0, then the mean of the labels is $\frac{2}{3}$ for this segment. We hypothesize that the means of labels for AF segments are larger than the means of labels for Normal segments, since there should be more 1 labels for AF segments. After conducting the T-test, we get a p-value of $8.27\times 10^{-67}$, and we conclude that representations with same labels are more likely to be close in the latent T-S feature space. An AF ECG segment and a Normal ECG segment, along with segments close to them in the T-S feature space, are shown in Figure \ref{fig:nearest}.

\subsection{Results of model interpretation}

\begin{figure}
    \centering
    \includegraphics[width=\linewidth]{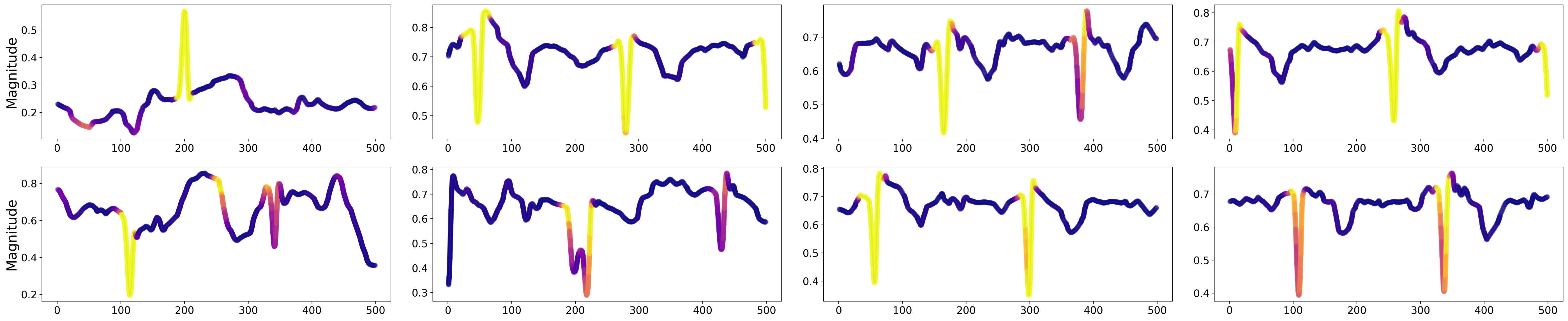}
    \includegraphics[width=\linewidth]{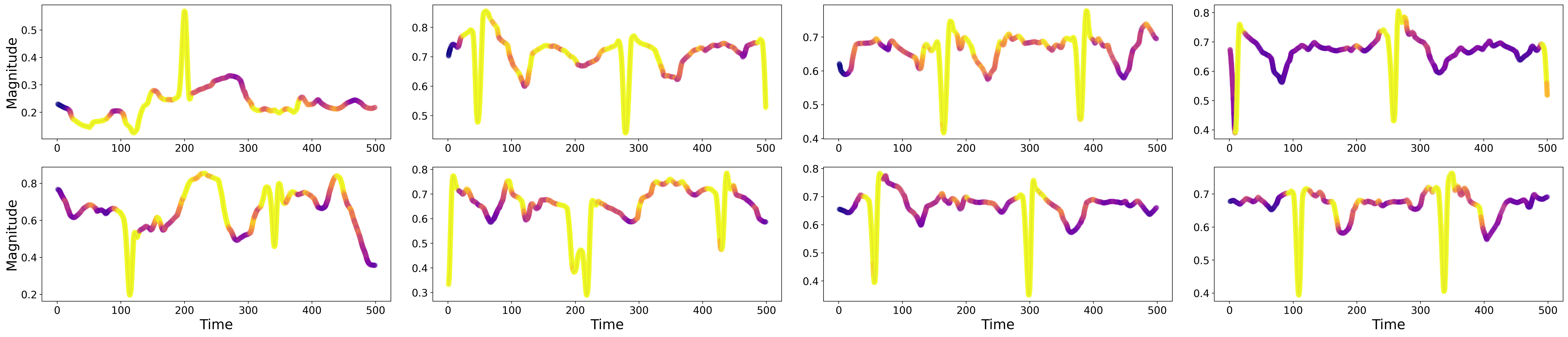}
    \includegraphics[width=.7\linewidth]{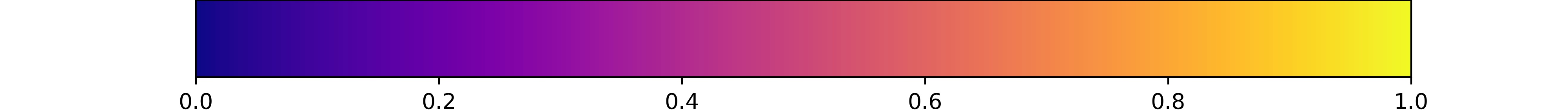}
    \caption{The LRP heatmaps on eight examples of ECG segments. The top two rows are from the model trained with spatial reverse detection, and the bottom two rows are from the model trained with temporal reverse detection. The areas with brighter colors have larger $R$ scores which contribute more to the model's final predictions. Temporal reverse helps the model pay attention to more details, while spatial reverse makes the model focus only on peaks.}
    \label{fig:lrp}
\end{figure}  
To interpret the experimental model, Figure \ref{fig:lrp} shows the LRP heatmaps. We see that the top two rows have a more balanced distribution of colors, which means that different parts of the ECG segments contribute more equally to the model trained with the temporal reverse detection task.
On the other hand, the bottom two rows correspond to the model trained with the spatial reverse detection task, and the non-peak regions are almost dark. It indicates that the relevance $R$ scores of these parts are extremely small, so the non-peak parts contribute little to the model's final predictions.

\section{Discussion}
Our detection of manipulated reverse signals can be easily performed in the real world without data labeling. Deep learning engineers can easily implement our method, which has comparable performance to sophisticated CL methods. 

In our experiment, we take AF detection as the downstream task. AF causes two distinct changes in ECG signals. First, the P-waves disappear with the appearance of rapid fibrillatory waves (F-waves) in the TQ intervals. Second, AF ECGs have absolutely irregular ventricular rhythms, resulting in unequal RR intervals. These two changes can correspond to our spatial reverse detection and temporal reverse detection respectively. The results shown in Table \ref{tab:results} indicate that detection of temporal reverse and spatial reverse both help improve the performance on downstream tasks. Spatial reverse detection might help the model recognize the R peaks well, and thus can detect the irregular ventricular rhythm which is related to the RR intervals. 

To evaluate the learned representations, we explore the data distributions in each feature space. As shown in Figure \ref{fig:tsne} and \ref{fig:nearest}, temporal reverse detection  and temporal-spatial reverse detection help the model project original ECG segments to a feature space where samples with same label are more likely to be closer. In contrast, spatial reverse detection shows little effect. To understand why temporal reverse detection is more effective than spatial reverse detection, we use LRP to illustrate how the model learns from original ECG segments under both tasks. The heatmaps of LRP in Figure \ref{fig:lrp} show that spatial reverse detection makes the model focus only on ECG peaks, which is in line with the intuition that the spatial reverse detection can be easily achieved by looking for ECG peaks. It implies that spatial reverse detection ignores some important spatial information such as P-waves. On the other hand, temporal reverse detection makes the model pay attention to most temporal points along ECG signals, which will help identify AF.

In future work, we plan to test our method on more downstream tasks to examine its generalization. Although our study is performed on a single-lead ECG dataset, the method can be easily extended to 12-lead ECG signals. Furthermore, we can apply the reverse paradigm to different methods. For example, we can construct negative pairs consisting of original signals and temporally reversed signals, and study its effectiveness in CL-based methods. 

\section{Conclusion}
In this paper, we propose a simple but effective self-supervised method to learn ECG representations via detecting manipulated reverses on unlabeled ECG data. The results show that our method outperforms other baseline methods on the AF detection downstream task. In addition, we visualize the data in the representation feature space and demonstrate that ECG segments with same class labels are close in the feature space, which indicates the learned ECG representations are effective on the downstream task. Ablation studies show that temporal reverse detection learns better representations compared to spatial reverse. With LRP interpretation, we find that temporal reverse detection enables the model to focus on more ECG locations, which may account for the better learning potential. Combining both temporal and spatial reverses, we provide an easy-to-implement and efficient method to learn ECG representations.

\section*{Acknowledgement}
This work was supported by the National Natural Science Foundation of China (No. 62102008).

\bibliographystyle{unsrtnat}
\bibliography{main}

\begin{thebibliography}{31}
\providecommand{\natexlab}[1]{#1}
\providecommand{\url}[1]{\texttt{#1}}
\expandafter\ifx\csname urlstyle\endcsname\relax
  \providecommand{\doi}[1]{doi: #1}\else
  \providecommand{\doi}{doi: \begingroup \urlstyle{rm}\Url}\fi

\bibitem[Hong et~al.(2020{\natexlab{a}})Hong, Zhou, Shang, Xiao, and
  Sun]{hong2020opportunities}
Shenda Hong, Yuxi Zhou, Junyuan Shang, Cao Xiao, and Jimeng Sun.
\newblock Opportunities and challenges of deep learning methods for
  electrocardiogram data: A systematic review.
\newblock \emph{Computers in Biology and Medicine}, 122:\penalty0 103801,
  2020{\natexlab{a}}.

\bibitem[Rahul and Sharma(2022)]{rahul2022artificial}
Jagdeep Rahul and Lakhan~Dev Sharma.
\newblock Artificial intelligence-based approach for atrial fibrillation
  detection using normalised and short-duration time-frequency ecg.
\newblock \emph{Biomedical Signal Processing and Control}, 71:\penalty0 103270,
  2022.

\bibitem[Han et~al.(2022)Han, Sun, Song, Zhao, Zhang, and
  Mao]{han2022detecting}
Jingyu Han, Guangpeng Sun, Xinhai Song, Jing Zhao, Jin Zhang, and Yi~Mao.
\newblock Detecting ecg abnormalities using an ensemble framework enhanced by
  bayesian belief network.
\newblock \emph{Biomedical Signal Processing and Control}, 72:\penalty0 103320,
  2022.
\newblock ISSN 1746-8094.
\newblock \doi{https://doi.org/10.1016/j.bspc.2021.103320}.
\newblock URL
  \url{https://www.sciencedirect.com/science/article/pii/S1746809421009174}.

\bibitem[Ertu{\u{g}}rul et~al.(2021)Ertu{\u{g}}rul, Acar, Aldemir, and
  {\"O}ztekin]{ertuugrul2021automatic}
{\"O}mer~Faruk Ertu{\u{g}}rul, Emrullah Acar, Erdo{\u{g}}an Aldemir, and
  Abdulkerim {\"O}ztekin.
\newblock Automatic diagnosis of cardiovascular disorders by sub images of the
  ecg signal using multi-feature extraction methods and randomized neural
  network.
\newblock \emph{Biomedical Signal Processing and Control}, 64:\penalty0 102260,
  2021.

\bibitem[Tong et~al.(2021)Tong, Sun, Zhou, Shen, Jiang, Sha, and
  Chang]{tong2021locating}
Yanni Tong, Yinan Sun, Peng Zhou, Yang Shen, Hua Jiang, Xianzheng Sha, and
  Shijie Chang.
\newblock Locating abnormal heartbeats in ecg segments based on deep weakly
  supervised learning.
\newblock \emph{Biomedical Signal Processing and Control}, 68:\penalty0 102674,
  2021.

\bibitem[Zhao et~al.(2021)Zhao, Xia, and Wang]{zhao2021dual}
Ranqi Zhao, Yi~Xia, and Qiuyang Wang.
\newblock Dual-modal and multi-scale deep neural networks for sleep staging
  using eeg and ecg signals.
\newblock \emph{Biomedical Signal Processing and Control}, 66:\penalty0 102455,
  2021.

\bibitem[Werth et~al.(2020)Werth, Radha, Andriessen, Aarts, and
  Long]{werth2020deep}
Jan Werth, Mustafa Radha, Peter Andriessen, Ronald~M Aarts, and Xi~Long.
\newblock Deep learning approach for ecg-based automatic sleep state
  classification in preterm infants.
\newblock \emph{Biomedical Signal Processing and Control}, 56:\penalty0 101663,
  2020.

\bibitem[Hong et~al.(2020{\natexlab{b}})Hong, Wang, and Fu]{hong2020cardioid}
Shenda Hong, Can Wang, and Zhaoji Fu.
\newblock Cardioid: Learning to identification from electrocardiogram data.
\newblock \emph{Neurocomputing}, 412:\penalty0 11--18, 2020{\natexlab{b}}.

\bibitem[Zhang et~al.(2021)Zhang, Zhao, Deng, Zhang, and Zhang]{zhang2021human}
Yefei Zhang, Zhidong Zhao, Yanjun Deng, Xiaohong Zhang, and Yu~Zhang.
\newblock Human identification driven by deep cnn and transfer learning based
  on multiview feature representations of ecg.
\newblock \emph{Biomedical Signal Processing and Control}, 68:\penalty0 102689,
  2021.
\newblock ISSN 1746-8094.
\newblock \doi{https://doi.org/10.1016/j.bspc.2021.102689}.
\newblock URL
  \url{https://www.sciencedirect.com/science/article/pii/S174680942100286X}.

\bibitem[Rasti-Meymandi and Ghaffari(2022)]{rasti2022deep}
Arash Rasti-Meymandi and Aboozar Ghaffari.
\newblock A deep learning-based framework for ecg signal denoising based on
  stacked cardiac cycle tensor.
\newblock \emph{Biomedical Signal Processing and Control}, 71:\penalty0 103275,
  2022.

\bibitem[Liu et~al.(2021)Liu, Zhao, and She]{liu2021self}
Han Liu, Zhenbo Zhao, and Qiang She.
\newblock Self-supervised ecg pre-training.
\newblock \emph{Biomedical Signal Processing and Control}, 70:\penalty0 103010,
  2021.

\bibitem[Sarkar and Etemad(2020)]{sarkar2020self}
Pritam Sarkar and Ali Etemad.
\newblock Self-supervised learning for ecg-based emotion recognition.
\newblock In \emph{ICASSP 2020-2020 IEEE International Conference on Acoustics,
  Speech and Signal Processing (ICASSP)}, pages 3217--3221. IEEE, 2020.

\bibitem[Kiyasseh et~al.(2021)Kiyasseh, Zhu, and Clifton]{kiyasseh2021clocs}
Dani Kiyasseh, Tingting Zhu, and David~A Clifton.
\newblock Clocs: Contrastive learning of cardiac signals across space, time,
  and patients.
\newblock In \emph{International Conference on Machine Learning}, pages
  5606--5615. PMLR, 2021.

\bibitem[Lan et~al.(2021)Lan, Ng, Hong, and Feng]{lan2021intra}
Xiang Lan, Dianwen Ng, Shenda Hong, and Mengling Feng.
\newblock Intra-inter subject self-supervised learning for multivariate cardiac
  signals.
\newblock \emph{arXiv preprint arXiv:2109.08908}, 2021.

\bibitem[Wold et~al.(1987)Wold, Esbensen, and Geladi]{wold1987principal}
Svante Wold, Kim Esbensen, and Paul Geladi.
\newblock Principal component analysis.
\newblock \emph{Chemometrics and intelligent laboratory systems}, 2\penalty0
  (1-3):\penalty0 37--52, 1987.

\bibitem[Bourlard and Kamp(1988)]{bourlard1988auto}
Herv{\'e} Bourlard and Yves Kamp.
\newblock Auto-association by multilayer perceptrons and singular value
  decomposition.
\newblock \emph{Biological cybernetics}, 59\penalty0 (4):\penalty0 291--294,
  1988.

\bibitem[Dasan and Panneerselvam(2021)]{dasan2021novel}
Evangelin Dasan and Ithayarani Panneerselvam.
\newblock A novel dimensionality reduction approach for ecg signal via
  convolutional denoising autoencoder with lstm.
\newblock \emph{Biomedical Signal Processing and Control}, 63:\penalty0 102225,
  2021.

\bibitem[Porumb et~al.(2020)Porumb, Griffen, Hattersley, and
  Pecchia]{porumb2020nocturnal}
Mihaela Porumb, Corbin Griffen, John Hattersley, and Leandro Pecchia.
\newblock Nocturnal low glucose detection in healthy elderly from one-lead ecg
  using convolutional denoising autoencoders.
\newblock \emph{Biomedical Signal Processing and Control}, 62:\penalty0 102054,
  2020.

\bibitem[Kuznetsov et~al.(2021)Kuznetsov, Moskalenko, Gribanov, and
  Zolotykh]{kuznetsov2021interpretable}
VV~Kuznetsov, VA~Moskalenko, DV~Gribanov, and Nikolai~Yu Zolotykh.
\newblock Interpretable feature generation in ecg using a variational
  autoencoder.
\newblock \emph{Frontiers in Genetics}, 12, 2021.

\bibitem[{Clifford} et~al.(2017){Clifford}, {Liu}, {Moody}, {Lehman}, {Silva},
  {Li}, {Johnson}, and {Mark}]{clifford2017af}
G.~D. {Clifford}, C.~{Liu}, B.~{Moody}, L.~H. {Lehman}, I.~{Silva}, Q.~{Li},
  A.~E. {Johnson}, and R.~G. {Mark}.
\newblock Af classification from a short single lead ecg recording: The
  physionet/computing in cardiology challenge 2017.
\newblock In \emph{2017 Computing in Cardiology (CinC)}, pages 1--4, 2017.
\newblock \doi{10.22489/CinC.2017.065-469}.

\bibitem[Xie et~al.(2017)Xie, Girshick, Dollár, Tu, and He]{ResNext}
Saining Xie, Ross Girshick, Piotr Dollár, Zhuowen Tu, and Kaiming He.
\newblock Aggregated residual transformations for deep neural networks.
\newblock In \emph{2017 IEEE Conference on Computer Vision and Pattern
  Recognition (CVPR)}, pages 5987--5995, 2017.
\newblock \doi{10.1109/CVPR.2017.634}.

\bibitem[Radosavovic et~al.(2020)Radosavovic, Kosaraju, Girshick, He, and
  Dollár]{radosavovic2020designing}
Ilija Radosavovic, Raj~Prateek Kosaraju, Ross Girshick, Kaiming He, and Piotr
  Dollár.
\newblock Designing network design spaces, 2020.

\bibitem[Hong et~al.(2019)Hong, Zhou, Wu, Shang, Wang, Li, and Xie]{Hong_2019}
Shenda Hong, Yuxi Zhou, Meng Wu, Junyuan Shang, Qingyun Wang, Hongyan Li, and
  Junqing Xie.
\newblock Combining deep neural networks and engineered features for cardiac
  arrhythmia detection from {ECG} recordings.
\newblock \emph{Physiological Measurement}, 40\penalty0 (5):\penalty0 054009,
  jun 2019.
\newblock \doi{10.1088/1361-6579/ab15a2}.

\bibitem[Hong et~al.(2020{\natexlab{c}})Hong, Xu, Khare, Priambada, Maher,
  Aljiffry, Sun, and Tumanov]{hong2020holmes}
Shenda Hong, Yanbo Xu, Alind Khare, Satria Priambada, Kevin Maher, Alaa
  Aljiffry, Jimeng Sun, and Alexey Tumanov.
\newblock Holmes: Health online model ensemble serving for deep learning models
  in intensive care units.
\newblock In \emph{Proceedings of the 26th ACM SIGKDD International Conference
  on Knowledge Discovery \& Data Mining}, pages 1614--1624, 2020{\natexlab{c}}.

\bibitem[Bach et~al.(2015)Bach, Binder, Montavon, Klauschen, M{\"u}ller, and
  Samek]{bach2015pixel}
Sebastian Bach, Alexander Binder, Gr{\'e}goire Montavon, Frederick Klauschen,
  Klaus-Robert M{\"u}ller, and Wojciech Samek.
\newblock On pixel-wise explanations for non-linear classifier decisions by
  layer-wise relevance propagation.
\newblock \emph{PloS one}, 10\penalty0 (7):\penalty0 e0130140, 2015.

\bibitem[Sturm et~al.(2016)Sturm, Lapuschkin, Samek, and
  M{\"u}ller]{sturm2016interpretable}
Irene Sturm, Sebastian Lapuschkin, Wojciech Samek, and Klaus-Robert M{\"u}ller.
\newblock Interpretable deep neural networks for single-trial eeg
  classification.
\newblock \emph{Journal of neuroscience methods}, 274:\penalty0 141--145, 2016.

\bibitem[Banluesombatkul et~al.(2020)Banluesombatkul, Ouppaphan, Leelaarporn,
  Lakhan, Chaitusaney, Jaimchariya, Chuangsuwanich, Chen, Phan, Dilokthanakul,
  et~al.]{banluesombatkul2020metasleeplearner}
Nannapas Banluesombatkul, Pichayoot Ouppaphan, Pitshaporn Leelaarporn,
  Payongkit Lakhan, Busarakum Chaitusaney, Nattapong Jaimchariya, Ekapol
  Chuangsuwanich, Wei Chen, Huy Phan, Nat Dilokthanakul, et~al.
\newblock Metasleeplearner: A pilot study on fast adaptation of
  bio-signals-based sleep stage classifier to new individual subject using
  meta-learning.
\newblock \emph{IEEE Journal of Biomedical and Health Informatics}, 2020.

\bibitem[He et~al.(2016)He, Zhang, Ren, and Sun]{he2016deep}
Kaiming He, Xiangyu Zhang, Shaoqing Ren, and Jian Sun.
\newblock Deep residual learning for image recognition.
\newblock In \emph{2016 IEEE Conference on Computer Vision and Pattern
  Recognition (CVPR)}, pages 770--778, 2016.
\newblock \doi{10.1109/CVPR.2016.90}.

\bibitem[Chen et~al.(2020)Chen, Kornblith, Norouzi, and Hinton]{chen2020simple}
Ting Chen, Simon Kornblith, Mohammad Norouzi, and Geoffrey Hinton.
\newblock A simple framework for contrastive learning of visual
  representations.
\newblock In \emph{International conference on machine learning}, pages
  1597--1607. PMLR, 2020.

\bibitem[Fawcett(2006)]{AUC}
Tom Fawcett.
\newblock An introduction to roc analysis.
\newblock \emph{Pattern Recogn. Lett.}, 27\penalty0 (8):\penalty0 861–874,
  June 2006.
\newblock ISSN 0167-8655.
\newblock \doi{10.1016/j.patrec.2005.10.010}.
\newblock URL \url{https://doi.org/10.1016/j.patrec.2005.10.010}.

\bibitem[Kubat et~al.(1998)Kubat, Holte, and Matwin]{gmean}
M.~Kubat, R.~C. Holte, and S.~Matwin.
\newblock Machine learning for the detection of oil spills in satellite radar
  images.
\newblock \emph{Machine Learning}, 30\penalty0 (2-3):\penalty0 195--215, 1998.

\end{thebibliography}

\end{document}